# Estimation of Markov Chain via Rank-Constrained Likelihood

Xudong Li [1]   Mengdi Wang [1]   Anru Zhang [2]

## Abstract

This paper studies the estimation of low-rank Markov chains from empirical trajectories. We propose a non-convex estimator based on rank-constrained likelihood maximization. Statistical upper bounds are provided for the Kullback-Leiber divergence and the $\ell_2$ risk between the estimator and the true transition matrix. The estimator reveals a compressed state space of the Markov chain. We also develop a novel DC (difference of convex function) programming algorithm to tackle the rank-constrained non-smooth optimization problem. Convergence results are established. Experiments show that the proposed estimator achieves better empirical performance than other popular approaches.

## 1. Introduction

In scientific studies, engineering systems, and social environments, one often has to interact with complex processes with uncertainties, collect data, and learn to make predictions and decisions. A critical question is how to model the unknown random process from data. To answer this question, we particularly focus on discrete Markov chains with a finite but potentially huge state space. Suppose we are given a sequence of observations generated by an unknown Markov process. The true transition matrix is unknown, and the physical states of the system and its law of transition is hidden under massive noisy observations. We are interested in the estimation and state compression of Markov processes, i.e., to identify compact representations of the state space and recover a reduced-dimensional Markov model.

To be more specific, we focus on the estimation of a class of Markov chains with low-rank transition matrices in this paper. Low-rankness is ubiquitous in practice. For example, the random walk of a taxi turns out to correspond to a nearly low-rank Markov chain. Each taxi trip can be viewed as a sample transition of a city-wide random walk that characterizes the traffic dynamics. For another example, control and reinforcement learning, which are typically modeled as Markov decision processes, suffer from the curse of dimensionality (Sutton & Barto, 1998). It is of vital interest to identify the compressed representation of the "state" of game, in order to reduce the dimensionality of policy and value functions. Multiple efforts show that as long as a reduced model or a compact state representation is given, one can solve high-dimensional control problems in sample-efficient ways; see (Singh et al., 1995; Van Roy, 2006; Malek et al., 2014) for examples.

The low-rank Markov chain is a basic model for analyzing data generated by stochastic processes. It is closely related to a variety of latent-variable models, including Markov process with rich observations (Azizzadenesheli et al., 2016), state aggregation models (Singh et al., 1995) and hidden Markov models (Hsu et al., 2012). Low-rankness of Markov chains can be used to recover network partition under some lumpability assumption (E et al., 2008). In particular, low-rank Markov chain can be viewed as a special form of hidden Markov model where the observations contain all the information of the hidden states.

In this article, we propose a rank-constrained maximum likelihood approach for transition matrix estimation from empirical trajectories. The statistical guarantees is developed for the proposed estimator. Especially, the upper bounds for the Kullback-Leibler (KL) divergence between the estimator and the true transition matrix as well as the $\ell_2$ risk are established. We further provide a minimax lower bound to show that the proposed estimator is near rate-optimal within a class of low-rank Markov chains. A related recent work (Zhang & Wang, 2017) considered the low-rank estimation by a spectral method and established upper bound for the total variation. Compared to the existing results, our maximum likelihood estimator and the KL-divergence bounds appear to be the first ones of this type.

The proposed estimator requires solving an optimization problem with both rank and polyhedral constraints. Due to the presence of non-convex rank constraint, the optimiza-tion problem is difficult – there is no efficient approach that directly applies to such a non-convex problem, to the best of our knowledge. In this paper, we use a penalty approach to relax the rank constraint and transform the original problem into a DC (difference of convex functions) programming one. Furthermore, we develop a corresponding DC algorithm to solve the problem. The algorithm proceeds by solving a sequence of inner subproblems, for which we develop an efficient subroutine based on multi-block alternating direction method of multipliers (ADMM). As a byproduct of this research, we develop a new class of DC algorithms and a unified convergence analysis for solving non-convex non-smooth problems.

---

[1]Department of Operations Research and Financial Engineering, Princeton University, Sherrerd Hall, Princeton, NJ 08544 [2]Department of Statistics, University of Wisconsin-Madison, Madison, WI 53706. Correspondence to: Xudong Li <xudongl@princeton.edu>, Mengdi Wang <mengdiw@princeton.edu>, Anru Zhang <anruzhang@stat.wisc.edu>.

Finally, the performance of the proposed estimator and algorithm is verified through simulation studies.

## 2. Related literatures

Model reduction for complicated systems has a long history that traces back to variable-resolution dynamic programming (Moore, 1991) and state aggregation for decision process (Sutton & Barto, 1998). In particular, (Deng et al., 2011; Deng & Huang, 2012) considered low-rank reduction of Markov models with explicitly known transition probability matrix, but not the estimation of the reduced models.

Low-rank models and spectral methods have been proved powerful in extracting features from high-dimensional data, with numerous applications including network analysis (E et al., 2008), community detection (Newman, 2013), ranking (Negahban et al., 2016), product recommendation (Keshavan et al., 2010) and many more. Estimation of hidden Markov models is related to spectral matrix/tensor decomposition, which has been considered in (Hsu et al., 2012). Later, (Anandkumar et al., 2014) studied a more general tensor-decomposition-based spectral approach, which can be applied to a variety of latent variable models including hidden Markov models. (Huang et al., 2016) considered the estimation of a rank-two probabilistic matrix from a few number of independent samples. Estimation of low-rank Markov chain was considered for the first time in (Zhang & Wang, 2017). Particularly, a spectral method via truncated singular value decomposition was introduced and the upper and lower error bounds in total variation were established. (Yang et al., 2017) developed an online stochastic gradient method for computing the leading singular space of a transition matrix from random walk data. The online stochastic approximation algorithms were also analyzed recently in (Li et al., 2017; 2018) for principal component estimation. The estimation of discrete distributions in the form of vectors is another line of related research, where the procedure and estimation risk has been considered in both classic and recent literature (Steinhaus, 1957; Lehmann & Casella, 2006; Han et al., 2015).

Form the perspective of optimization, DC programming is a backbone in handling the rank-constrained problem. First introduced by (Pham Dinh & Le Thi, 1997), the DC algorithm has become a prominent tool for handling a class of nonconvex optimization problems (see also (Pham Dinh & Le Thi, 2005; Le Thi et al., 2012; 2017; Le Thi & Pham Dinh, 2018)). In particular, (Van Dinh et al., 2015; Wen et al., 2017) considered the majorized indefinite-proximal DC algorithm, which is closely related to the optimization methods in this paper. However, both (Van Dinh et al., 2015; Wen et al., 2017) used the majorization technique with restricted choices of majorants, neither considered the introduction of the indefinite proximal terms, and (Wen et al., 2017) further requires the smooth part in the objective to be convex. A more flexible algorithmic framework and a unified convergence analysis is provided in Section 5.4.

## 3. Low-rank Markov chains and latent variable models

Consider a discrete-time Markov chain on $p$ states $\{s_1, \ldots, s_p\}$ with transition matrix $\mathbf{P} \in \Re^{p \times p}$. Then $\mathbf{P}$ satisfies $P_{ij} = P(s_j|s_i)$, $0 \leq P_{ij} \leq 1$, and $\sum_{j=1}^{p} P_{ij} = 1$ for $1 \leq i, j \leq p$. Let $\mu$ be the corresponding stationary distribution. Suppose $\text{rank}(\mathbf{P}) = r$. Motivated by the examples that we will discuss below, we focus on the case where $r \ll p$, i.e., the rank of $\mathbf{P}$ is significantly smaller than the total number of states.

The first example of low-rank Markov chain in reduced-order models is the state aggregation – a basic model for describing complicated systems (Bertsekas, 1995; Bertsekas & Tsitsiklis, 1995). If a process admits an inherent state aggregation structure, its states can be clustered into a small number of disjoint subsets, such that states from the same cluster have identical transition distribution. We say that a Markov chain is state aggregatable if there exists a partition $\mathcal{S}_1, \mathcal{S}_2, \ldots, \mathcal{S}_r$ such that

$$P(\cdot \mid s) = P(\cdot \mid s'), \qquad \forall s, s' \in \mathcal{S}_i, \ i \in [r].$$

It can be shown that the state aggregation yields a special factorization of transition matrix. Particularly, there exists $\mathbf{U}, \mathbf{V} \in \Re^{p \times r}$ such that

$$\mathbf{P} = \mathbf{U}\mathbf{V}^\top,$$

where each row of $\mathbf{U}$ has all 0's except exact one 1 and $\mathbf{V}$ is nonnegative.

Another related model is referred to as *Markov decision process with rich observations* (Azizzadenesheli et al., 2016) in the context of reinforcement learning. It is essentially a latent-variable model for Markov chains. We say that a Markov chain is a $r$-latent-variable model if there exists a stochastic process $\{z_t\} \subset [r]$ such that

$$P(z_t \mid s_t) = P(z_t \mid s_1, \ldots, s_t)$$

$$P(s_{t+1} \mid z_t) = P(s_{t+1} \mid s_1, \ldots, s_t, z_t).$$

The latent-variable model, another instance of low-rank Markov chains, is slightly more general than the state aggregation model. One can show that the transition matrix $\mathbf{P}$ of a $r$-latent-variable Markov chain has a nonnegative rank at most $r$ and there exists $\mathbf{U}, \mathbf{V} \in \Re^{p \times r}, \tilde{\mathbf{P}} \in \Re^{r \times r}$ such that

$$\mathbf{P} = \mathbf{U}\tilde{\mathbf{P}}\mathbf{V}^\top,$$

where $\tilde{\mathbf{P}}$ is a stochastic matrix, rows of $\mathbf{U}$ and columns of $\mathbf{V}$ are vectors of probability distributions.

A third related notion is the "lumpability" of Markov chain (Buchholz, 1994). We say that a Markov chain is lumpable with respect to a partition $\mathcal{S}_1, \mathcal{S}_2, \ldots, \mathcal{S}_r$ if for any $i, j \in [r]$ and $s', s \in \mathcal{S}_i$,

$$P(\mathcal{S}_j \mid s) = P(\mathcal{S}_j \mid s').$$

Lumpability characterizes some partition of the state space that preserves the strong Markov property. It implies that some eigenvectors of the transition matrix is equal to the indicator functions of the subsets. However, it is a weaker assumption than state aggregation and does not necessarily imply low-rankness. As pointed out by (E et al., 2008), the lumpable partition can be recovered by clustering the singular vectors of the transition matrix.

The readers are also referred to (Zhang & Wang, 2017) for more discussions on low-rank Markov process examples. In summary, low-rankness is a ubiquitous in a broad class of reduced-order Markov models. Estimation of low-rank transition matrices involves estimating the leading subspace, which provides a venue towards state compression.

## 4. Minimax rate-optimal estimation of low-rank Markov chains

Now we consider the estimation of transition matrix based on a single state-transition trajectory. Given a Markov chain $\mathbf{X} = \{X_0, X_1, \ldots, X_n\}$, we aim to estimate the transition matrix $\mathbf{P}$ given the assumption that $\text{rank}(\mathbf{P}) = r$ and $r \ll p$. For $1 \leq i, j \leq p$, denote the transition counts from state $s_i$ to $s_j$ by $n_{ij}$:

$$n_{ij} := |\{1 \leq k \leq N \mid X_{k-1} = s_i, X_k = s_j\}|.$$

Let $n_i := \sum_{j=1}^p n_{ij}$ for $i = 1, \ldots, p$ and $n = \sum_{i=1}^p n_i$.

**Proposition 1.** *The negative log-likelihood of $\mathbf{P}$ based on state-transition trajectory $\{x_0, \ldots, x_n\}$ is*

$$L(\mathbf{P}) := -\frac{1}{n} \sum_{i=1}^p \sum_{j=1}^p n_{ij} \log(P_{ij}). \tag{1}$$

We propose the following rank-constrained maximum likelihood estimator:

$$\begin{aligned} \widehat{\mathbf{P}} = & \arg\min L(\mathbf{Q}) \\ \text{s.t.} \quad & \mathbf{Q}\mathbf{1}_p = \mathbf{1}_p, \ \text{rank}(\mathbf{Q}) \leq r, \\ & 0 \leq Q_{ij} \leq 1, \quad \forall 1 \leq i, j \leq p. \end{aligned} \tag{2}$$

Then we aim to analyze the theoretical property of $\widehat{\mathbf{P}}$. Since the direct analysis of (2) for general transition matrices is technical challenging, we consider a slightly different setting, assume $\mathbf{P}$ is entry-wise bounded, and analyze the rank-constrained optimization with corresponding constraints. As we will show later in numerical analysis, such additional assumptions and constraints may not be necessary in practice.

**Theorem 1** (Statistical Recovery Guarantee). *Suppose the transition matrix $\mathbf{P}$ satisfies $\text{rank}(\mathbf{P}) \leq r$, $\alpha/p \leq \mathbf{P}_{ij} \leq \beta/p$ for constants $0 < \alpha < 1 < \beta$. Consider the following programming,*

$$\begin{aligned} \widehat{\mathbf{P}} = & \arg\min L(\mathbf{Q}) \\ \text{s.t.} \quad & \mathbf{Q}\mathbf{1}_p = \mathbf{1}_p, \ \text{rank}(\mathbf{Q}) \leq r, \\ & \alpha \leq p Q_{ij} \leq \beta \ \ \forall 1 \leq i, j \leq p. \end{aligned} \tag{3}$$

*If we observe a total number of $n$ independent state transitions generated from the stationary distribution and $n \geq Cp \log(p)$, the optimal solution $\widehat{\mathbf{P}}$ of problem (3) satisfies*

$$D_{KL}(\mathbf{P}, \widehat{\mathbf{P}}) = \sum_{i,j=1}^p \mu_i P_{ij} \log(P_{ij}/\widehat{P}_{ij})$$

$$\leq C \left( \frac{pr \log p}{n} + \sqrt{\frac{\log p}{n}} \right) \tag{4}$$

$$\sum_{i=1}^p \|P_{i\cdot} - \widehat{P}_{i\cdot}\|_2^2 \leq C \left( \frac{pr \log p}{n} + \sqrt{\frac{\log p}{n}} \right) \tag{5}$$

*with probability at least $1 - Cp^{-c}$. Here, $C, c > 0$ are universal constants.*

The proof is given in Section B of the supplementary material.

**Remark 1.** Note that the result in Theorem 1 applies to the data when each state transition is sampled independently from the stationary distribution of the Markov chain. In practice, samples from Markov chains are typically dependent. Then we can artificially introduce independence by sampling the dependent random walk data. Specifically, let $\tau(\epsilon)$ be the $\epsilon$-mixing time of the Markov chain (see, e.g., (Levin & Peres, 2017)). We pick a sufficiently small $\epsilon$ and sample one transition every $\tau(\epsilon)$ transitions so that the down-sampled data are nearly independent (whose distribution is within $\epsilon$ total-variation distance from the independent data). Note that $\tau(\epsilon) \approx \tau(1/4) \log(1/\epsilon)$ scales logarithmic in $1/\epsilon$. So we can pick $\epsilon$ to be sufficiently small without deteriorating the analysis much. Therefore, the overall sample complexity in the case of random walk data is roughly $\tau(1/4)$ times the sample complexity in the independent case (up to polylog factors).

**Remark 2.** In the proof of Theorem 1, in order to bound the difference between the empirical and real loss functions for the rank-constraint estimator, we apply the concentration inequalities for empirical process and a "peeling scheme." In particular, we consider a partition for the set of all possible $\widehat{\mathbf{P}}$, then derive estimation loss upper bounds for each of these subsets based on concentration inequalities. The techniques used here are related to recent works on matrix completion (Negahban & Wainwright, 2012), although our problem setting, methodology, and sampling scheme are all very different from matrix completion.

**Remark 3.** In Theorem 1, we use entry-wise boundedness condition $\alpha$ & $\beta$ instead of the incoherence condition, to ensure bounded Hessian of the KL-divergence. $\alpha$ & $\beta$ do not appear in bounds as they are treated as constants – dependence on $\alpha$ & $\beta$ is complicated and beyond the scope of this paper. Similar condition of entry-wise boundedness was used in matrix completion (Recht, 2011, Assumption A1) and discrete distribution estimation (Kamath et al., 2015). In fact, based on (Kamath et al., 2015, Theorem 10), the entry-wise lower bound is necessary for deriving reasonable statistical recovery bounds in KL-divergence.

Define the following set of transition matrices $\mathcal{P} = \{\mathbf{P} : \mathbf{P}1_p = 1_p, 0 \leq \mathbf{P}_{ij} \leq 1, \operatorname{rank}(\mathbf{P}) \leq r\}$. We further have the following lower bound results.

**Theorem 2** (Lower Bound). *When $n \geq Cpr$, $\widehat{\mathbf{P}}$ is any estimator for the transition matrix $\mathbf{P}$ based on a sample trajectory of length $n$. Then,*

$$\inf_{\widehat{\mathbf{P}}} \sup_{\mathbf{P} \in \mathcal{P}} \mathbb{E} D_{KL}(\mathbf{P}, \widehat{\mathbf{P}}) \geq c \frac{pr}{n};$$

$$\inf_{\widehat{\mathbf{P}}} \sup_{\mathbf{P} \in \mathcal{P}} \mathbb{E} \sum_{i=1}^{p} \|\widehat{P}_{i\cdot} - P_{i\cdot}\|_2^2 \geq c \frac{pr}{n},$$

*where $C, c$ are universal constants.*

Its proof is given in Section C of the supplementary material. Theorems 1 and 2 together show that the proposed estimator is rate-optimal up to a logarithm factor, as long as $n = O(pr \log p)$.

**Remark 4.** Especially when $r = 1$, $\mathbf{P}$ can be written as $1v^\top$ for some vector $v \in \Re^p$, and then estimating $\mathbf{P}$ essentially reduces to estimating a discrete distribution from multinomial count data. Our upper bound in Theorem 1 nearly matches (up to a log factor) the classical results of discrete distribution estimation $\ell_2$ risks (see, e.g. (Lehmann & Casella, 2006, Pg. 349)).

In addition to the full transition matrix $\mathbf{P}$, the leading left and right singular subspaces of $\mathbf{P}$, say $\mathbf{U}, \mathbf{V} \in \Re^{p \times r}$, also play important roles in Markov chain analysis. For example, by performing $k$-means on the reliable estimations of $\mathbf{U}$ or $\mathbf{V}$ for a state aggregatable Markov chains, one can achieve good performance of state aggregation (Zhang & Wang, 2017). Based on previous discussions, one can further establish the error bounds on singular subspace estimation for Markov transition matrix. The proof of the following theorem is given in Section D of the supplementary material.

**Theorem 3.** *Under the setting of Theorem 1, suppose the left and right singular subspaces of $\widehat{\mathbf{P}}$ are $\widehat{\mathbf{U}} \in \Re^{p \times r}$ and $\widehat{\mathbf{V}} \in \Re^{p \times r}$, where $\widehat{\mathbf{P}}$ is the optimizer of (3), one has*

$$\max\{\|\sin\Theta(\widehat{\mathbf{U}}, \mathbf{U})\|_F^2, \|\sin\Theta(\widehat{\mathbf{V}}, \mathbf{V})\|_F^2\}$$
$$\leq \min\left\{\frac{C(pr\log p)/n + C\sqrt{(\log p)/n}}{\sigma_r^2(\mathbf{P})}, r\right\}$$

*with probability at least $1 - Cp^{-c}$. Here, $\sigma_r(\mathbf{P})$ is the $r$-th singular value of $\mathbf{P}$ and $\|\sin\Theta(\widehat{\mathbf{U}}, \mathbf{U})\|_F = (r - \|\widehat{\mathbf{U}}^\top \mathbf{U}\|_F^2)^{1/2}$.*

**Remark 5.** Similarly as Theorem 2, one can develop the corresponding lower bound to show the optimality for the upper bound in Theorem 3.

## 5. Optimization methods for the rank-constrained likelihood problem

In this section, we develop the optimization method for computing the rank-constrained likelihood maximizer (2). In Section 5.1, a penalty approach is applied to transform the original intractable rank-constrained problem into a DC programming problem. Then we solve this problem by a proximal DC (PDC) algorithm in Section 5.2. We discuss the solver for the subproblems involved in the proximal DC algorithm in Section 5.3. Lastly, a unified convergence analysis of a class of majorized indefinite-proximal DC (Majorized iPDC) algorithms is provided in Section 5.4.

### 5.1. A penalty approach for problem (2)

Recall (2) is intractable due to the non-convex rank constraint, we introduce a penalty approach to relax such a constraint, and particularly study the following general optimization problem:

$$\min\{f(\mathbf{X}) \mid \mathcal{A}(\mathbf{X}) = b, \operatorname{rank}(\mathbf{X}) \leq r\}, \quad (6)$$

where $f : \Re^{p \times p} \to (-\infty, +\infty]$ is a closed, convex, but possibly non-smooth function, $\mathcal{A} : \Re^{p \times p} \to \Re^m$ is a linear map, $b \in \Re^m$ and $r > 0$ are given data. Especially when $f(\mathbf{X}) = -\frac{1}{n}\sum_{i=1}^{p}\sum_{j=1}^{p} n_{ij} \log(X_{ij}) + \delta(\mathbf{X} \mid \mathbf{X} \geq 0)$, $\mathcal{A}(\mathbf{X}) = \mathbf{X}1_p$, $b = 1_p$, and $\delta(\cdot \mid \mathbf{X} \geq 0)$ is the indicator function of the closed convex set $\{\mathbf{X} \in \Re^{p \times p} \mid X \geq 0\}$, the general problem (6) becomes the original rank-constraint maximum likelihood problem (2).

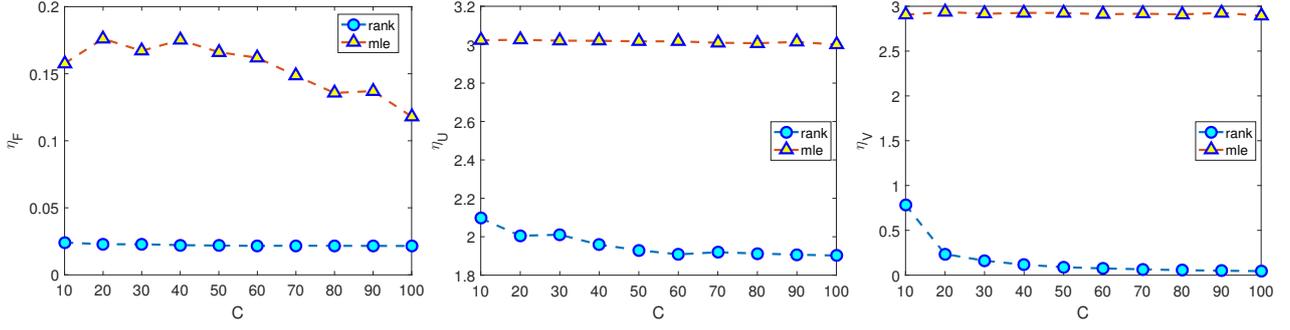

*Figure 1.* Comparison between **rank** and **mle** with accuracy measures $(\eta_F, \eta_U, \eta_V)$ versus number of state jumps $n = C^2 rp \log(p)$.

Given $\mathbf{X} \in \Re^{p \times p}$, let $\sigma_1(\mathbf{X}) \geq \cdots \geq \sigma_p(\mathbf{X}) \geq 0$ be the singular values of $\mathbf{X}$. Since $\text{rank}(\mathbf{X}) \leq r$ if and only $\sigma_{r+1}(\mathbf{X}) + \ldots + \sigma_p(\mathbf{X}) = \|\mathbf{X}\|_* - \|\mathbf{X}\|_{(r)} = 0$, where $\|\mathbf{X}\|_{(r)} = \sum_{i=1}^{r} \sigma_i(\mathbf{X})$ is the Ky Fan $r$-norm of $\mathbf{X}$, (6) can be equivalently formulated as

$$\min \left\{ f(\mathbf{X}) \mid \|\mathbf{X}\|_* - \|\mathbf{X}\|_{(r)} = 0, \mathcal{A}(\mathbf{X}) = b \right\}.$$

See also (Gao & Sun, 2010, equation (29)). The penalized formulation of problem (6) is

$$\min_{\mathbf{X} \in \Re^{p \times p}} \left\{ f(\mathbf{X}) + c(\|\mathbf{X}\|_* - \|\mathbf{X}\|_{(r)}) \mid \mathcal{A}(\mathbf{X}) = b \right\}, \quad (7)$$

where $c > 0$ is a penalty parameter. Since $\|\cdot\|_{(r)}$ is convex, it is clear that the objective in problem (7) is a difference of two convex functions: $f(\mathbf{X}) + c\|\mathbf{X}\|_*$ and $c\|\mathbf{X}\|_{(r)}$, i.e., (7) is a DC program.

Let $\mathbf{X}_c^*$ be an optimal solution to the penalized problem (7). The following proposition shows that $\mathbf{X}_c^*$ is also the optimizer to (6) when it is low-rank.

**Proposition 2.** *If* $\text{rank}(\mathbf{X}_c^*) \leq r$, *then* $\mathbf{X}_c^*$ *is also an optimal solution to the original problem* (6).

In practice, one can gradually increase the penalty parameter $c$ to obtain a sufficient low rank solution $\mathbf{X}_c^*$. In our numerical experiments, we can obtain solutions with the desired rank with a properly chosen parameter $c$.

### 5.2. A PDC algorithm for penalized problem (7)

The central idea of the DC algorithm (Pham Dinh & Le Thi, 1997) is as follows: at each iteration, one approximates the concave part of the objective function by its affine majorant, then solves the resulting convex optimization problem. In this subsection, we present a variant of the classic DC algorithm for solving (7). For the execution of the algorithm, we recall that the sub-gradient of Ky Fan $r$-norm at a point $\mathbf{X} \in \Re^{p \times p}$ (Watson, 1993) is

$$\partial \|\mathbf{X}\|_{(r)} = \left\{ \mathbf{U} \, \text{Diag}(q^*) \mathbf{V}^T \mid q^* \in \Delta \right\},$$

where $\mathbf{U}$ and $\mathbf{V}$ are the singular vectors of $\mathbf{X}$, $\Delta$ is the optimal solution set of the following problem

$$\max_{q \in \Re^p} \left\{ \sum_{i=1}^{p} \sigma_i(\mathbf{X}) q_i \mid \langle \mathbf{1}_p, q \rangle = r, 0 \leq q \leq 1 \right\}.$$

Note that one can efficiently obtain a component of $\partial \|\mathbf{X}\|_{(r)}$ by computing the SVD and picking up the SVD vectors corresponding to the $r$ largest singular values. After these preparations, we are ready to state the PDC algorithm for problem (7). Different from the classic DC algorithm, an additional proximal term is added to ensure the existence of solutions of subproblems (8) and the convergence of the difference of two consecutive iterations generated by the algorithm. See Theorem 4 and Remark 6 for more details.

---

**Algorithm 1** A PDC algorithm for solving problem (7)

Given $c > 0$, $\alpha \geq 0$, and the stopping tolerance $\eta$, choose initial point $\mathbf{X}^0 \in \Re^{p \times p}$. Iterate the following steps for $k = 0, 1, \ldots$:

**1.** Choose $\mathbf{W}_k \in \partial \|\mathbf{X}^k\|_{(r)}$. Compute

$$\mathbf{X}^{k+1} = \arg\min \left\{ \begin{array}{l} f(\mathbf{X}) + c(\|\mathbf{X}\|_* - \langle \mathbf{W}_k, \mathbf{X} - \mathbf{X}^k \rangle \\ - \|\mathbf{X}^k\|_{(r)}) + \frac{\alpha}{2} \|\mathbf{X} - \mathbf{X}^k\|_F^2 \\ \mid \mathcal{A}(\mathbf{X}) = b \end{array} \right\}. \quad (8)$$

**2.** If $\|\mathbf{X}^{k+1} - \mathbf{X}^k\|_F \leq \eta$, stop.

---

We say that $\mathbf{X}$ is a critical point of problem (7) if

$$\partial(f(\mathbf{X}) + c\|\mathbf{X}\|_*) \cap (c\partial \|\mathbf{X}\|_{(r)}) \neq \emptyset.$$

We state the following convergence results for Algorithm 1.

**Theorem 4.** *Let* $\{\mathbf{X}^k\}$ *be the sequence generated by Algorithm 1 and* $\alpha \geq 0$. *Then* $\{f(\mathbf{X}^k) + c(\|\mathbf{X}^k\|_* - \|\mathbf{X}^k\|_{(r)})\}$ *is a non-increasing sequence. If* $\mathbf{X}^{k+1} = \mathbf{X}^k$ *for some integer* $k \geq 0$, *then* $\mathbf{X}^k$ *is a critical point of* (7). *Otherwise,*

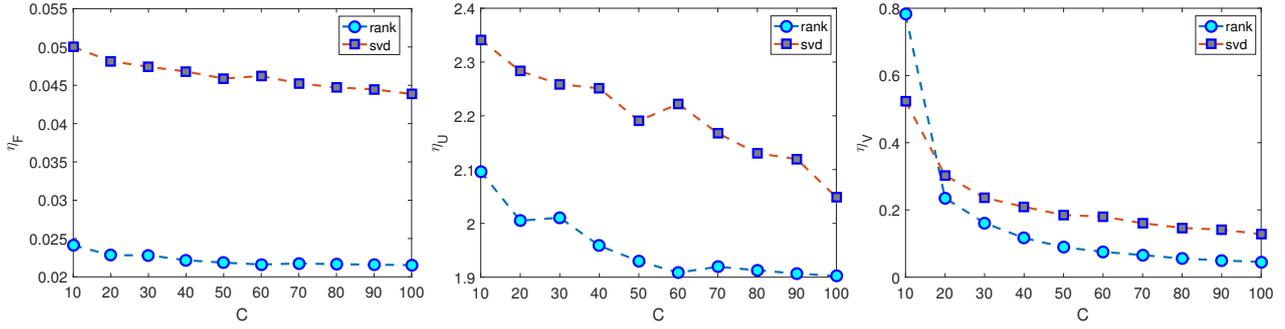

*Figure 2.* Comparison between **rank** and **svd** with accuracy measures $(\eta_F, \eta_U, \eta_V)$ VS. number of state jumps $n = C^2 rp \log(p)$.

*it holds that*

$$\begin{aligned}&\left(f(\mathbf{X}^{k+1}) + c(\|\mathbf{X}^{k+1}\|_* - \|\mathbf{X}^{k+1}\|_{(r)})\right) \\ &\quad - \left(f(\mathbf{X}^k) + c(\|\mathbf{X}^k\|_* - \|\mathbf{X}^k\|_{(r)})\right) \\ &\leq -\frac{\alpha}{2}\|\mathbf{X}^{k+1} - \mathbf{X}^k\|_F^2.\end{aligned}$$

*Moreover, any accumulation point of the bounded sequence $\{\mathbf{X}^k\}$ is a critical point of problem (7). In addition, if $\alpha > 0$, it holds that $\lim_{k\to\infty}\|\mathbf{X}^{k+1} - \mathbf{X}^k\|_F = 0$.*

**Remark 6** (Selection of Parameters). In practice, a small $\alpha > 0$ is suggested to ensure strict decrease of the objective value and convergence of $\{\|\mathbf{X}^{k+1} - \mathbf{X}^k\|_F\}$; if $f$ is strongly convex, one achieves these nice properties even if $\alpha = 0$ based on the results of Theorem 6. The penalty parameter $c$ can be adaptively adjusted according to the rank of the sequence generated by Algorithm 1.

### 5.3. A multi-block ADMM for subproblem (8)

It is important and non-trivial to solve the convex subproblem (8) in Algorithm 1. Note that (8) is a nuclear norm penalized convex optimization problem, we propose to apply an efficient multi-block alternating direction method of multipliers (ADMM) for solving the dual of (8). A comprehensive numerical study has been conducted in (Li et al., 2016b) and justifies our procedure.

Instead of working directly on (8), we study a slightly general model as follows. Given $\mathbf{W} \in \Re^{p\times p}$ and $\alpha \geq 0$, consider

$$(\mathbf{P}) \quad \begin{aligned}\min \quad & f(\mathbf{X}) + \langle \mathbf{W}, \mathbf{X}\rangle + c\|\mathbf{X}\|_* + \frac{\alpha}{2}\|\mathbf{X}\|_F^2 \\ \text{s.t.} \quad & \mathcal{A}(\mathbf{X}) = b.\end{aligned}$$

Its dual problem can be written as

$$(\mathbf{D}) \quad \begin{aligned}\min \quad & f^*(-\Xi) - \langle b, y\rangle + \frac{\alpha}{2}\|\mathbf{Z}\|_F^2 \\ \text{s.t.} \quad & \Xi + \mathcal{A}^*(y) + \mathbf{S} + \alpha\mathbf{Z} = \mathbf{W}, \quad \|\mathbf{S}\|_2 \leq c,\end{aligned}$$

where $f^*$ is the conjugate function of $f$, $\|\cdot\|_2$ denotes the spectral norm. When we set $f$ to be the negative likelihood function (1), $f^*$ becomes

$$\begin{aligned}f^*(\Xi) = & \sum_{(i,j)\in\Omega}\frac{n_{ij}}{n}(\log\frac{n_{ij}}{n} - 1 - \log(-\Xi_{ij})) \\ & + \sum_{(i,j)\in\overline{\Omega}}\delta(\Xi_{ij} \mid \Xi_{ij} \leq 0) \quad \forall\,\Xi \in \Re^{p\times p},\end{aligned}$$

where $\Omega = \{(i,j) \mid n_{ij} \neq 0\}$ and $\overline{\Omega} = \{(i,j) \mid n_{ij} = 0\}$. Given $\sigma > 0$, the augmented Lagrangian function $\mathcal{L}_\sigma$ associated with (**D**) is

$$\mathcal{L}_\sigma(\Xi, y, \mathbf{S}, \mathbf{Z}; \mathbf{X}) = f^*(-\Xi) - \langle b, y\rangle + \frac{\alpha}{2}\|\mathbf{Z}\|_F^2$$
$$+ \frac{\sigma}{2}\|\Xi + \mathcal{A}^*(y) + \mathbf{S} + \alpha\mathbf{Z} - \mathbf{W} + \mathbf{X}/\sigma\|^2 - \frac{1}{2\sigma}\|\mathbf{X}\|^2.$$

We consider ADMM type methods for solving problem (**D**). Since there are four separable blocks in (**D**) (namely $\Xi$, $y$, $\mathbf{S}$, and $\mathbf{Z}$), the direct extended ADMM is not applicable. Fortunately, the functions corresponding to blocks $y$ and $\mathbf{Z}$ in the objective of (**D**) are linear-quadratic. Thus we can apply the multi-block symmetric Gauss-Sediel based ADMM (sGS-ADMM) (Li et al., 2016b). In literature (Chen et al., 2017; Ferreira et al., 2017; Lam et al., 2018; Li et al., 2016b; Wang & Zou, 2018), extensive numerical experiments demonstrate that sGS-ADMM is not only convergent but also faster than the directly extended multi-block ADMM and its many other variants. Specifically, the algorithmic framework of sGS-ADMM for solving (**D**) is presented in Algorithm 2. Note that when $\alpha = 0$, the computation steps corresponding to block $\mathbf{Z}$ will not be performed.

When implementing Algorithm 2, only partial SVD, which is much cheaper than full SVD, is needed as $r \ll p$. The following convergence results can be directly obtained from (Li et al., 2016b). A sketch of the proof is given in supplementary material.

**Theorem 5.** *Suppose that the solution sets of (**P**) and (**D**) are nonempty. Let $\{(\Xi^k, y^k, \mathbf{S}^k, \mathbf{Z}^k, \mathbf{X}^k)\}$ be the sequence generated by Algorithm 2. If $\tau \in (0, (1+\sqrt{5})/2)$, then the sequence $\{(\Xi^k, y^k, \mathbf{S}^k, \mathbf{Z}^k)\}$ converges to an optimal solution of (**D**) and $\{\mathbf{X}^k\}$ converges to an optimal solution of (**P**).*

**Algorithm 2** An sGS-ADMM for solving (**D**)

**Input:** initial point $(\boldsymbol{\Xi}^0, y^0, \mathbf{S}^0, \mathbf{Z}^0, \mathbf{X}^0)$, penalty parameter $\sigma > 0$, maximum iteration number $K$, and the steplength $\gamma \in (0, (1+\sqrt{5})/2)$
**for** $k = 0$ **to** $K$ **do**
  $y^{k+\frac{1}{2}} = \arg\min_y \mathcal{L}_\sigma(\boldsymbol{\Xi}^k, y, \mathbf{S}^k, \mathbf{Z}^k; \mathbf{X}^k)$
  $\boldsymbol{\Xi}^{k+1} = \arg\min_{\boldsymbol{\Xi}} \mathcal{L}_\sigma(\boldsymbol{\Xi}, y^{k+\frac{1}{2}}, \mathbf{S}^k, \mathbf{Z}^k; \mathbf{X}^k)$
  $y^{k+1} = \arg\min_y \mathcal{L}_\sigma(\boldsymbol{\Xi}^{k+1}, y, \mathbf{S}^k, \mathbf{Z}^k; \mathbf{X}^k)$
  $\mathbf{Z}^{k+\frac{1}{2}} = \arg\min_{\mathbf{Z}} \mathcal{L}_\sigma(\boldsymbol{\Xi}^{k+1}, y^{k+1}, \mathbf{S}^k, \mathbf{Z}; \mathbf{X}^k)$
  $\mathbf{S}^{k+1} = \arg\min_{\mathbf{S}} \mathcal{L}_\sigma(\boldsymbol{\Xi}^{k+1}, y^{k+1}, \mathbf{S}, \mathbf{Z}^{k+\frac{1}{2}}; \mathbf{X}^k)$
  $\mathbf{Z}^{k+1} = \arg\min_{\mathbf{Z}} \mathcal{L}_\sigma(\boldsymbol{\Xi}^{k+1}, y^{k+1}, \mathbf{S}^{k+1}, \mathbf{Z}; \mathbf{X}^k)$
  $\mathbf{X}^{k+1} = \mathbf{X}^k + \gamma\sigma(\boldsymbol{\Xi}^{k+1} + \mathcal{A}^*(y^{k+1}) + \mathbf{S}^{k+1} + \alpha \mathbf{Z}^{k+1} - \mathbf{W})$
**end for**

### 5.4. A unified analysis for the majorized iPDC algorithm

Due to the presence of the proximal term $\frac{\alpha}{2}\|\mathbf{X} - \mathbf{X}^k\|^2$ in Algorithm 1, the classical DC analyses cannot be applied directly. Hence, in this subsection, we provide a unified convergence analysis for the majorized indefinite-proximal DC (majorized iPDC) algorithm which includes Algorithm 1 as a special instance. Let $\mathbb{X}$ be a finite-dimensional real Euclidean space endowed with inner product $\langle \cdot, \cdot \rangle$ and induced norm $\|\cdot\|$. Consider the following optimization problem

$$\min_{x \in \mathbb{X}} \theta(x) \triangleq g(x) + p(x) - q(x), \quad (9)$$

where $g : \mathbb{X} \to \Re$ is a continuously differentiable function (not necessarily convex) with a Lipschitz continuous gradient and Lipschitz modulus $L_g > 0$, i.e.,

$$\|\nabla f(x) - \nabla f(x')\| \le L_g \|x - x'\| \quad \forall\, x, x' \in \mathbb{X},$$

$p : \mathbb{X} \to (-\infty, +\infty]$ and $q : \mathbb{X} \to (-\infty, +\infty]$ are two proper closed convex functions. It is not difficult to observe that penalized problem (7) is a special instance of problem (9). For general model (9), one can only expect the DC algorithm converges to a critical point $\bar{x} \in \mathbb{X}$ of (9) satisfying

$$(\nabla g(\bar{x}) + \partial p(\bar{x})) \cap \partial q(\bar{x}) \neq \emptyset.$$

Since $g$ is continuously differentiable with Lipschitz continuous gradient, there exists a self-adjoint positive semidefinite linear operator $\mathcal{G} : \mathbb{X} \to \mathbb{X}$ such that for any $x, x' \in \mathbb{X}$,

$$g(x) \le \widehat{g}(x; x') \triangleq g(x') + \langle \nabla g(x'), x - x' \rangle + \frac{1}{2}\|x - x'\|^2_{\mathcal{G}}.$$

The majorized iPDC algorithm for solving (9) is presented in Algorithm 3. We further provide the following convergence results.

**Algorithm 3** A majorized indefinite-proximal DC algorithm for solving problem (9)

Given initial point $x^0 \in \mathbb{X}$ and stopping tolerance $\eta$, choose a self-adjoint linear operator $\mathcal{T} : \mathbb{X} \to \mathbb{X}$ and $\mathcal{G}$. Iterate the following steps for $k = 0, 1, \dots$:
**1.** Choose $\xi^k \in \partial q(x^k)$. Compute

$$x^{k+1} \in \arg\min_{x \in \mathbb{X}} \left\{ \widehat{\theta}(x; x^k) + \frac{1}{2}\|x - x^k\|^2_{\mathcal{T}} \right\}, \quad (10)$$

where $\widehat{\theta}(x; x^k) \triangleq \widehat{g}(x; x^k) + p(x) - \big(q(x^k) + \langle x - x^k, \xi^k \rangle\big)$.
**2.** If $\|x^{k+1} - x^k\| \le \eta$, stop.

**Theorem 6.** *Assume that $\inf_{x \in \mathbb{X}} \theta(x) > -\infty$. Let $\{x^k\}$ be the sequence generated by Algorithm 3. If $x^{k+1} = x^k$ for some $k \ge 0$, then $x^k$ is a critical point of (9). If $\mathcal{G} + 2\mathcal{T} \succeq 0$, then any accumulation point of $\{x^k\}$, if exists, is a critical point of (9). In addition, if $\mathcal{G} + 2\mathcal{T} \succ 0$, it holds that $\lim_{i \to \infty} \|x^{k+1} - x^k\| = 0$.*

The proof of Theorem 6 and more discussions are provided in the supplementary material.

## 6. Simulation Results

We first generate the rank-$r$ transition matrix as $\mathbf{P} = \mathbf{U}\boldsymbol{\Sigma}\mathbf{V}^T$, where $\mathbf{U}, \mathbf{V} \in \Re^{p \times r}$ have orthonormal columns and the diagonal matrix $\boldsymbol{\Sigma} = \text{diag}(\sigma_i) \in \Re^{r \times r}$ with $\sigma_i > 0$ for $i = 1, \dots, r$. Then we simulate a Markov chain trajectory of length $n = C^2 r p \log(p)$. Here, $r = 10$, $p = 500$ and $C$ is a varying constant from $10$ to $10^2$.

We compare four estimation procedures, i.e., maximum likelihood estimator (**mle**), truncated SVD estimator (**svd**) (Zhang & Wang, 2017), nuclear norm penalized estimator (**nu**), and our rank-constrained maximum likelihood estimator (**rank**). For each estimator $\widehat{\mathbf{P}}$, let $\widehat{\mathbf{U}}$ and $\widehat{\mathbf{V}}$ be the leading $r$ left and right singular vectors of $\widehat{\mathbf{P}}$, respectively. We measure the performance of $\widehat{\mathbf{P}}$ through three quantities:

$$\eta_F = \|\mathbf{P} - \widehat{\mathbf{P}}\|_F / \sqrt{p}, \quad \eta_U = \|\sin\Theta(\mathbf{U}, \widehat{\mathbf{U}})\|_F,$$
$$\eta_V = \|\sin\Theta(\mathbf{V}, \widehat{\mathbf{V}})\|_F.$$

The comparison between **mle** and **rank** is presented in Figure 1. As one can observe, the empirical error of our rank-constrained maximum likelihood estimator is significantly smaller than the plain maximum likelihood estimator.

From Figure 2, one can see that our rank-constrained estimator outperforms **svd** in terms of all three accuracy measurements. Recently, (Zhang & Wang, 2017) studied the **svd** approach and developed the total variation error bounds. The rank-constrained estimator and the KL-divergence error bounds studied in this paper are and harder to analyze and more meaningful for discrete distribution estimation.

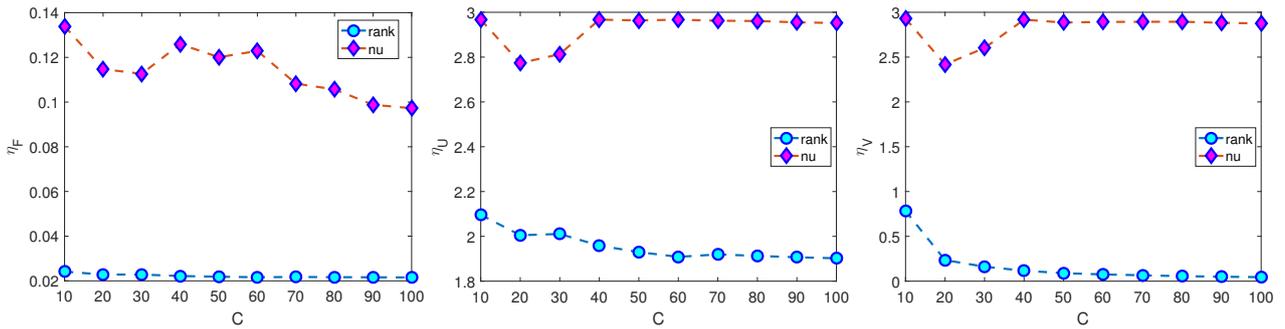

*Figure 3.* Comparison between **rank** and **nu** with accuracy measures $(\eta_F, \eta_U, \eta_V)$ VS. number of state jumps $n = C^2 rp \log(p)$.

In the implementation of nuclear norm penalized estimator **nu**, the penalty parameter is chosen via 5-fold cross-validation. The comparison between **rank** and **nu** is plotted in Figure 3. Clearly, the estimation error of **rank** is much smaller than that of **nu**. In fact, as one can see from Algorithm 1, the nuclear norm penalized estimator is in fact a single step in the procedure of our rank-constrained estimator, and **rank** gradually refines based on **nu**. Theoretically speaking, one could expect that, similar to the matrix recovery (Recht et al., 2010), the rank-constrained estimator enjoys better statistical performances and needs weaker assumption over the nuclear norm penalized estimator. Moreover, Theorems 1 and 2 indicate that the rank-constrained maximum likelihood estimator is statistically optimal for the estimation of Markov Chains. However, such a result is not clear for the nuclear norm approach.

## 7. Conclusion

This paper studies the recovery and state compression of low-rank Markov chains from empirical trajectories via a rank-constrained likelihood approach. We provide statistical upper bounds for the $\ell_2$ risk and Kullback-Leiber divergence between the estimator and the true probability transition matrix for the proposed estimator. Then, a novel DC programming algorithm is developed to solve the associated rank-constrained optimization problem. The proposed algorithm non-trivially combines several recent optimization techniques, such as the penalty approach, the proximal DC algorithm, and the multi-block sGS-ADMM. We further study a new class of majorized indefinite-proximal DC algorithms for solving general non-convex non-smooth DC programming problems and provide a unified convergence analysis. Experiments on simulated data illustrate the merits of our approach.

## A. Proof of Proposition 1

*Proof.* Given $x_k = i$, $x_{k+1}$ is with discrete distribution $\mathbf{P}_{i\cdot}$. Thus, the log-likelihood of $x_{k+1}|x_k = \log(P_{x_k,x_{k+1}}) = \langle \log(\mathbf{P}), e_{x_k}e_{x_{k+1}}^\top \rangle$. Then the negative log-likelihood given $\{x_0, \ldots, x_n\}$ is

$$-\sum_{k=1}^n \log(P_{x_k,x_{k+1}}) = -\sum_{k=1}^n \langle \log(\mathbf{P}), e_{x_k}e_{x_{k+1}}^\top \rangle = -\sum_{i=1}^p \sum_{j=1}^p n_{ij} \log(P_{ij}).$$

□

## B. Proof of Theorem 1

*Proof.* Recall $D_{KL}(\mathbf{P}, \mathbf{Q}) = \sum_{i=1}^p \mu_i D_{KL}(P_{i\cdot}, Q_{i\cdot}) = \sum_{j=1}^p \mu_i P_{ij} \log(P_{ij}/Q_{ij})$. For convenience, we also denote,

$$\tilde{D}(\mathbf{P}, \mathbf{Q}) = \frac{1}{n} \sum_{k=1}^n \langle \log(\mathbf{P}) - \log(\mathbf{Q}), \mathbf{E}_k \rangle,$$

where $\mathbf{E}_k = e_i e_j^\top$ if the $k$-th jump is from States $i$ to $j$. Then $(\mathbf{E}_k)_{k=1}^n$ be independent copies such that $P(\mathbf{E}_k = e_i e_j^\top) = \mu_i P_{ij}$, and

$$L(\mathbf{P}) = -\frac{1}{n} \sum_{i,j=1}^p n_{ij} \log(P_{ij}) = -\frac{1}{n} \sum_{k=1}^n \log\langle \mathbf{X}, \mathbf{E}_k \rangle$$

By the property of the programming,

$$\tilde{D}(\mathbf{P}, \widehat{\mathbf{P}}) = \frac{1}{n} \sum_{k=1}^n \langle \log(\mathbf{P}) - \log(\widehat{\mathbf{P}}), \mathbf{E}_k \rangle = L(\widehat{\mathbf{P}}) - L(\mathbf{P}) \leq 0. \tag{11}$$

Based on the assumption, $\text{rank}(\mathbf{P}) \wedge \text{rank}(\widehat{\mathbf{P}}) \leq r$. For any $\mathbf{Q}$ with $\text{rank}(\mathbf{Q}) \leq r$, we must have $\text{rank}(\mathbf{Q} - \mathbf{P}) \leq 2r$. Due to the duality between operator and spectral norm,

$$\|\mathbf{Q} - \mathbf{P}\|_* \leq \sqrt{2r}\|\mathbf{Q} - \mathbf{P}\|_F. \tag{12}$$

Next, we denote $\eta = C_\eta \sqrt{\log p/n}$ for some large constant $C_\eta > 0$, and introduce the following deterministic set in $\mathbb{R}^{p \times p}$,

$$\mathcal{C} = \{\mathbf{Q} : \alpha/p \leq Q_{ij} \leq \beta/p, \ \text{rank}(Q) \leq r, D_{KL}(\mathbf{P}, \mathbf{Q}) \geq \eta\}.$$

We particularly aim to show next that

$$P\left\{\forall \mathbf{Q} \in \mathcal{C}, \ \left|\tilde{D}(\mathbf{P}, \mathbf{Q}) - D_{KL}(\mathbf{P}, \mathbf{Q})\right| \leq \frac{1}{2} D_{KL}(\mathbf{P}, \mathbf{Q}) + \frac{Cpr \log(p)}{n}\right\} \geq 1 - Cp^{-c}. \tag{13}$$

In order to prove (13), we first split $\mathcal{C}$ as the union of the sets,

$$\mathcal{C}_l = \{\mathbf{Q} \in \mathcal{C} : 2^{l-1}\eta \leq D_{KL}(\mathbf{P}, \mathbf{Q}) \leq 2^l \eta, \ \text{rank}(Q) \leq r\}, \quad l = 1, 2, 3, \ldots.$$

where $\eta$ is to be determined later. Define

$$\gamma_l = \sup_{\mathbf{Q} \in \mathcal{C}_l} \left|D_{KL}(\mathbf{P}, \mathbf{Q}) - \tilde{D}(\mathbf{P}, \mathbf{Q})\right|$$

$$= \sup_{\mathbf{Q} \in \mathcal{C}_l} \left|\frac{1}{n} \sum_{k=1}^n \langle \log(\mathbf{P}) - \log(\mathbf{Q}), \mathbf{E}_k \rangle - \mathbb{E}\langle \log(\mathbf{P}) - \log(\mathbf{Q}), \mathbf{E}_k \rangle\right|.$$

Since $|\log(P_{ij}) - \log(Q_{ij})| \leq \log(\beta/\alpha)$, we apply a empirical process version of Hoeffding's inequality (Theorem 14.2 in (Bühlmann & Van De Geer, 2011)),

$$P\left(\gamma_l - \mathbb{E}(\gamma_l) \geq 2^{l-3} \cdot \eta\right) \leq \exp\left(-\frac{cn \cdot 4^{l-3}\eta^2}{(\log(\beta/\alpha))^2}\right). \tag{14}$$

Estimation of Markov Chain via Rank-constrained Likelihoodfor constant $c > 0$. We generate $\{\varepsilon_k\}_{k=1}^n$ as i.i.d. Rademacher random variables. By a symmetrization argument in empirical process,

$$\mathbb{E}\gamma_l = \mathbb{E}\left(\sup_{\mathbf{Q}\in\mathcal{C}_l}\left|\frac{1}{n}\sum_{k=1}^n\langle\log\mathbf{P}-\log\mathbf{Q},\mathbf{E}_k\rangle - \mathbb{E}\frac{1}{n}\sum_{k=1}^n\langle\log\mathbf{P}-\log\mathbf{Q},\mathbf{E}_k\rangle\right|\right)$$
$$\leq 2\mathbb{E}\left(\sup_{\mathbf{Q}\in\mathcal{C}_l}\left|\frac{1}{n}\sum_{k=1}^n\varepsilon_k\langle\log\mathbf{P}-\log\mathbf{Q},\mathbf{E}_k\rangle\right|\right).$$

Let $\phi_k(t) = \alpha/p \cdot \langle\log(\mathbf{P})-\log(\mathbf{Q}+t),\mathbf{E}_k\rangle$, then $\phi_k(0)=0$ and $|\phi_k'(t)|\leq 1$ for all $t$ if $t+P_{ij}\geq \alpha/p$. In other words, $\phi_{k,i,j}$ is a contraction map for $t\geq \min_{i,j}(P_{ij}-\alpha/p)$. By concentration principle (Theorem 4.12 in (Ledoux & Talagrand, 2013)),

$$\mathbb{E}(\gamma_l) \leq \frac{2p}{\alpha}\mathbb{E}\left(\sup_{\mathbf{Q}\in\mathcal{C}_l}\left|\frac{1}{n}\sum_{k=1}^n\varepsilon_k\phi_k\left(\langle\mathbf{Q}-\mathbf{P},\mathbf{E}_k\rangle\right)\right|\right)$$
$$\leq \frac{4p}{\alpha}\mathbb{E}\left(\sup_{\mathbf{Q}\in\mathcal{C}_l}\left|\frac{1}{n}\sum_{k=1}^n\varepsilon_k\langle\mathbf{Q}-\mathbf{P},\mathbf{E}_k\rangle\right|\right)$$
$$\leq \frac{4p}{\alpha}\mathbb{E}\left(\sup_{\mathbf{Q}\in\mathcal{C}_l}\left\|\frac{1}{n}\sum_{k=1}^n\varepsilon_k\mathbf{E}_k\right\|\cdot\|\mathbf{Q}-\mathbf{P}\|_*\right)$$
$$\leq \frac{4p}{\alpha}\mathbb{E}\left\|\frac{1}{n}\sum_{k=1}^n\varepsilon_k\mathbf{E}_k\right\|\cdot\sup_{\mathbf{Q}\in\mathcal{C}_l}\|\mathbf{Q}-\mathbf{P}\|_* \quad (15)$$

By $\mathrm{rank}(\mathbf{P})\wedge\mathrm{rank}(\mathbf{Q})\leq r$ and Lemma 5 in (Zhang & Wang, 2017),

$$\sup_{\mathbf{Q}\in\mathcal{C}_l}\|\mathbf{Q}-\mathbf{P}\|_* \overset{(12)}{\leq} \sup_{\mathbf{Q}\in\mathcal{C}_l}\sqrt{2r}\|\mathbf{Q}-\mathbf{P}\|_F$$
$$\leq \sqrt{\frac{r(\beta/p)^2}{(\alpha/p)}\sum_{i=1}^p D(P_{i\cdot},Q_{i\cdot})} \leq \sqrt{\frac{r\beta^2}{\alpha^2}\cdot 2^l\eta}. \quad (16)$$

Then we evaluate $\mathbb{E}\|\frac{1}{n}\sum_{k=1}^n\varepsilon_k\mathbf{E}_k\|$. Note that $\|\mathbf{E}_k\|\leq 1$,

$$\|\sum_{k=1}^n\mathbb{E}\mathbf{E}_k^\top\mathbf{E}_k\| = n\left\|\sum_{i=1}^p\sum_{j=1}^p\mu_i P_{ij}(e_ie_j^\top)^\top(e_ie_j^\top)\right\| = n\left\|\sum_{j=1}^p(\mu^\top P)_j e_je_j^\top\right\|$$
$$= n\left\|\sum_{j=1}^p\mu_j e_je_j^\top\right\|\leq n\mu_{\max};$$

$$\|\sum_{k=1}^n\mathbb{E}\mathbf{E}_k\mathbf{E}_k^\top\| = n\left\|\sum_{i=1}^p\sum_{j=1}^p\mu_i P_{ij}(e_ie_j^\top)(e_ie_j^\top)^\top\right\| = \left\|\sum_{i=1}^p\sum_{j=1}^p\mu_i P_{ij}e_ie_i^\top\right\|$$
$$= \left\|\sum_{j=1}^p\mu_j e_je_j^\top\right\|\leq n\mu_{\max}.$$

By Theorem 1 in (Tropp, 2016),

$$\mathbb{E}\left\|\frac{1}{n}\sum_{k=1}^n\varepsilon_k\mathbf{E}_k\right\| \leq \frac{C\sqrt{n\mu_{\max}\log p}}{n}+\frac{C\log p}{n}\leq C\sqrt{\frac{\mu_{\max}\log p}{n}}\leq \sqrt{\frac{\beta\log p}{np}}. \quad (17)$$

provided that $n \geq Cp\log(p)$. Combining (14), (15), (16), and (17), we have

$$\mathbb{E}\gamma_l \leq C\sqrt{\frac{pr\log p}{n} \cdot 2^l \eta} \leq C^2 \frac{pr\log p}{2n} + 2^{l-3}\eta,$$

$$P\left(\gamma_l \geq 2^{l-2}\eta + \frac{Cpr\log p}{n}\right) \leq \exp\left(-cn \cdot 4^l \eta^2\right).$$

Now,

$$P\left(\exists \mathbf{Q} \in \mathcal{C}, \ \left|\tilde{D}(\mathbf{P},\mathbf{Q}) - D_{KL}(\mathbf{P},\mathbf{Q})\right| > \frac{1}{2}D_{KL}(\mathbf{P},\mathbf{Q}) + \frac{Cpr\log(p)}{n}\right)$$

$$\leq \sum_{l=0}^{\infty} P\left(\exists \mathbf{Q} \in \mathcal{C}_l, \ \left|\tilde{D}(\mathbf{P},\mathbf{Q}) - D_{KL}(\mathbf{P},\mathbf{Q})\right| > \frac{1}{2}D_{KL}(\mathbf{P},\mathbf{Q}) + \frac{Cpr\log(p)}{n}\right)$$

$$\leq \sum_{l=0}^{\infty} P\left(\exists \mathbf{Q} \in \mathcal{C}_l, \ \gamma_l > 2^{l-2}\eta + \frac{Cpr\log(p)}{n}\right)$$

$$\leq \sum_{l=0}^{\infty} \exp(-c \cdot C_\eta \cdot 4^l \log p) \leq \exp(-c \cdot C_\eta l \log(p)) \leq Cp^{-c}$$

provided reasonably large $C_\eta > 0$. Thus, we have obtained (13).

Finally, it remains to bound the errors for $\|\widehat{\mathbf{P}} - \mathbf{P}\|_F$ and $D_{KL}(\mathbf{P}, \widehat{\mathbf{P}})$ given (13). In fact, provided that (13) holds,

- if $\widehat{\mathbf{P}} \notin \mathcal{C}$, we have $D_{KL}(\mathbf{P}, \widehat{\mathbf{P}}) \leq C\sqrt{\frac{\log p}{n}}$;

- if $\widehat{\mathbf{P}} \in \mathcal{C}$, by (13),
$$D_{KL}(\mathbf{P}, \widehat{\mathbf{P}}) \leq \tilde{D}(\mathbf{P}, \widehat{\mathbf{P}}) + \frac{Cpr\log p}{n} \overset{(11)}{\leq} \frac{Cpr\log p}{n}.$$

To sum up, we must have

$$D_{KL}(\mathbf{P}, \widehat{\mathbf{P}}) \leq C\sqrt{\frac{\log p}{n}} + \frac{Cpr\log p}{n}.$$

with probability at least $1 - Cp^{-c}$. For Frobenius norm error, we shall note that

$$\|\widehat{\mathbf{P}} - \mathbf{P}\|_F^2 \leq \sum_{i=1}^{p} \|P_{i\cdot} - \widehat{P}_{i\cdot}\|_2^2 \leq \sum_{i=1}^{p} \frac{2\beta^2}{\alpha p} D_{KL}(P_{i\cdot}, \widehat{P}_{i\cdot})$$

$$\leq \sum_{i=1}^{p} \frac{2\beta^2}{\alpha^2} \mu_i D_{KL}(P_{i\cdot}, \widehat{P}_{i\cdot}) = \frac{\beta^2}{\alpha^2} D_{KL}(\mathbf{P}, \widehat{\mathbf{P}}).$$

Therefore, we have finished the proof for Theorem 1. □

## C. Proof of Theorem 2

*Proof.* Based on the proof of Theorem 1 in (Zhang & Wang, 2017), one has

$$\inf_{\widehat{\mathbf{P}}} \sup_{\mathbf{P} \in \bar{\mathcal{P}}} \frac{1}{p} \sum_{i=1}^{p} \mathbb{E}\|\widehat{P}_{i\cdot} - P_{i\cdot}\|_1 \geq c\left(\sqrt{\frac{rp}{n}} \wedge 1\right),$$

where $\bar{\mathcal{P}} = \{\mathbf{P} \in \mathcal{P} : 1/(2p) \leq P_{ij} \leq 3/(2p)\} \subseteq \mathcal{P}$. By Cauchy Schwarz inequality,

$$\sum_{i=1}^{p} \|\widehat{P}_{i\cdot} - P_{i\cdot}\|_1 = \sum_{i,j=1}^{p} |\widehat{P}_{ij} - P_{ij}| \leq p\sqrt{\sum_{i,j=1}^{p} (\widehat{P}_{ij} - P_{ij})^2},$$

Thus,

$$\inf_{\widehat{\mathbf{P}}} \sup_{\mathbf{P} \in \mathcal{P}} \mathbb{E} \sum_{i=1}^{p} \|\widehat{P}_{i\cdot} - P_{i\cdot}\|_2^2 \geq \left( \inf_{\widehat{\mathbf{P}}} \sup_{\mathbf{P} \in \mathcal{P}} \mathbb{E} \sum_{i=1}^{p} \frac{1}{p} \|\widehat{P}_{i\cdot} - P_{i\cdot}\|_1 \right)^2 \geq c\left(\frac{rp}{n} \wedge 1\right) \geq \frac{cpr}{n}.$$

The lower bound for KL divergence essentially follows due to the inequalities between $\ell_2$ and KL-divergence for bounded vectors in Lemma 5 of (Zhang & Wang, 2017). □

## D. Proof of Theorem 3

*Proof.* Let $\widehat{\mathbf{U}}_\perp, \widehat{\mathbf{V}}_\perp \in \Re^{p \times (p-r)}$ be the orthogonal complement of $\widehat{\mathbf{U}}$ and $\widehat{\mathbf{V}}$. Since $\mathbf{U}, \mathbf{V}, \widehat{\mathbf{U}}$, and $\widehat{\mathbf{V}}$ are the leading left and right singular vectors of $\mathbf{P}$ and $\widehat{\mathbf{P}}$, we have

$$\|\widehat{\mathbf{P}} - \mathbf{P}\|_F \geq \|\widehat{\mathbf{U}}_\perp^\top (\widehat{\mathbf{P}} - \mathbf{U}\mathbf{U}^\top \mathbf{P})\|_F = \|\widehat{\mathbf{U}}_\perp^\top \mathbf{U}\mathbf{U}^\top \mathbf{P}\|_F \geq \|\widehat{\mathbf{U}}_\perp^\top \mathbf{U}\|_F \cdot \sigma_r(\mathbf{U}^\top \mathbf{P}) = \|\sin\Theta(\widehat{\mathbf{U}}, \mathbf{U})\|_F \cdot \sigma_r(\mathbf{P}).$$

Similar argument also applies to $\|\sin\Theta(\widehat{\mathbf{V}}, \mathbf{V})\|$. Thus,

$$\max\{\|\sin\Theta(\widehat{\mathbf{U}}, \mathbf{U})\|_F, \|\sin\Theta(\widehat{\mathbf{V}}, \mathbf{V})\|_F\} \leq \min\left\{\frac{\|\widehat{\mathbf{P}} - \mathbf{P}\|_F}{\sigma_r(\mathbf{P})}, \sqrt{r}\right\}.$$

The rest of the proof immediately follows from Theorem 1. □

## E. Proof of Proposition 2

*Proof.* Since $\text{rank}(\mathbf{X}_c^*) \leq r$, we know that $\mathbf{X}_c^*$ is in fact a feasible solution to the original problem (5) and $\|\mathbf{X}_c^*\|_* - \|\mathbf{X}_c^*\|_{(r)} = 0$. Therefore, for any feasible solution $X$ to (5), it holds that

$$f(\mathbf{X}_c^*) = f(\mathbf{X}_c^*) + c(\|\mathbf{X}_c^*\|_* - \|\mathbf{X}_c^*\|_{(r)})$$
$$\leq f(\mathbf{X}) + c(\|\mathbf{X}\|_* - \|\mathbf{X}\|_{(r)}) = f(\mathbf{X}).$$

This completes the proof of the proposition. □

## F. Proof of Theorem 5 (Convergence of sGS-ADMM)

*Proof.* In order to use (Li et al., 2016b, Theorem 3), we need to write problem (**D**) as following

$$\begin{aligned} \min \quad & f^*(-\boldsymbol{\Xi}) - \langle b, y \rangle + \delta(\mathbf{S} \mid \|\mathbf{S}\|_2 \leq c) + \tfrac{\alpha}{2}\|\mathbf{Z}\|_F^2 \\ \text{s.t.} \quad & \mathcal{F}(\boldsymbol{\Xi}) + \mathcal{A}_1^*(y) + \mathcal{G}(\mathbf{S}) + \mathcal{B}_1^*(\mathbf{Z}) = \mathbf{W}, \end{aligned}$$

where $\mathcal{F}, \mathcal{A}_1, \mathcal{G}$ and $\mathcal{B}_1$ are linear operators such that for all $(\boldsymbol{\Xi}, y, \mathbf{S}, \mathbf{Z}) \in \Re^{p \times p} \times \Re^n \times \Re^{p \times p} \times \Re^{p \times p}$, $\mathcal{F}(\boldsymbol{\Xi}) = \boldsymbol{\Xi}$, $\mathcal{A}_1^*(y) = \mathcal{A}^*(y)$, $\mathcal{G}(\mathbf{S}) = \mathbf{S}$ and $\mathcal{B}_1^*(\mathbf{Z}) = \alpha \mathbf{Z}$. Clearly, $\mathcal{F} = \mathcal{G} = \mathcal{I}$ and $\mathcal{B}_1 = \alpha \mathcal{I}$ where $\mathcal{I} : \Re^{p \times p} \to \Re^{p \times p}$ is the identity map. Therefore, we have $\mathcal{A}_1 \mathcal{A}_1^* \succ 0$ and $\mathcal{F}\mathcal{F}^* = \mathcal{G}\mathcal{G}^* = \mathcal{I} \succ 0$. Note that if $\alpha > 0$, $\mathcal{B}_1 \mathcal{B}_1^* = \alpha^2 \mathcal{I} \succ 0$. Hence, the assumptions and conditions in (Li et al., 2016b, Theorem 3) are satisfied whenever $\alpha \geq 0$. The convergence results thus follow directly. □

## G. Proof of Theorems 4 and 6

We only need to prove Theorem 6 as Theorem 4 is a special incidence. To prove Theorem 6, we first introduce the following lemma.

**Lemma 1.** *Suppose that $\{x^k\}$ is the sequence generated by Algorithm 3. Then $\theta(x^{k+1}) \leq \theta(x^k) - \frac{1}{2}\|x^{k+1} - x^k\|_{\mathcal{G}+2\mathcal{T}}^2$.*

*Proof.* For any $k \geq 0$, by the optimality condition of problem (10) at $x^{k+1}$, we know that there exist $\eta^{k+1} \in \partial p(x^{k+1})$ such that

$$0 = \nabla g(x^k) + (\mathcal{G} + \mathcal{T})(x^{k+1} - x^k) + \eta^{k+1} - \xi^k = 0.$$

Then for any $k \geq 0$, we deduce

$$\theta(x^{k+1}) - \theta(x^k) \leq \widehat{\theta}(x^{k+1}; x^k) - \theta(x^k)$$
$$= p(x^{k+1}) - p(x^k) + \langle x^{k+1} - x^k, \nabla g(x^k) - \xi^k \rangle + \tfrac{1}{2}\|x^{k+1} - x^k\|^2_{\mathcal{G}}$$
$$\leq \langle \nabla g(x^k) + \eta^{k+1} - \xi^k, x^{k+1} - x^k \rangle + \tfrac{1}{2}\|x^{k+1} - x^k\|^2_{\mathcal{G}}$$
$$= -\tfrac{1}{2}\|x^{k+1} - x^k\|^2_{\mathcal{G}+2\mathcal{T}}.$$

This completes the proof of this lemma. $\square$

Now we are ready to prove Theorem 6.

*Proof.* From the optimality condition at $x^{k+1}$, we have that

$$0 \in \nabla g(x^k) + (\mathcal{G} + \mathcal{T})(x^{k+1} - x^k) + \partial p(x^{k+1}) - \xi^k.$$

Since $x^{k+1} = x^k$, this implies that

$$0 \in \nabla g(x^k) + \partial p(x^k) - \partial q(x^k),$$

i.e., $x^k$ is a critical point. Observe that the sequence $\{\theta(x^k)\}$ is non-increasing since

$$\theta(x^{k+1}) \leq \widehat{\theta}(x^{k+1}; x^k) \leq \widehat{\theta}(x^k; x^k) = \theta(x^k), \quad k \geq 0.$$

Suppose that there exists a subsequence $\{x^{k_j}\}$ that converging to $\bar{x}$, i.e., one of the accumulation points of $\{x^k\}$. By Lemma 1 and the assumption that $\mathcal{G} + 2\mathcal{T} \succeq 0$, we know that for all $x \in \mathbb{X}$

$$\widehat{\theta}(x^{k_j+1}; x^{k_j+1}) = \theta(x^{k_j+1})$$
$$\leq \theta(x^{k_j+1}) \leq \widehat{\theta}(x^{k_j+1}; x^{k_j}) \leq \widehat{\theta}(x; x^{k_j}).$$

By letting $j \to \infty$ in the above inequality, we obtain that

$$\widehat{\theta}(\bar{x}; \bar{x}) \leq \widehat{\theta}(x; \bar{x}).$$

By the optimality condition of $\widehat{\theta}(x; \bar{x})$, we have that there exists $\bar{u} \in \partial p(\bar{x})$ and $\bar{v} \in \partial q(\bar{x})$ such that

$$0 \in \nabla g(\bar{x}) + \bar{u} - \bar{v}$$

This implies that $(\nabla g(\bar{x}) + \partial p(\bar{x})) \cap \partial q(\bar{x}) \neq \emptyset$. To establish the rest of this proposition, we obtain from Lemma 1 that

$$\lim_{t \to +\infty} \frac{1}{2} \sum_{i=0}^{t} \|x^{k+1} - x^k\|^2_{\mathcal{G}+2\mathcal{T}}$$
$$\leq \liminf_{t \to +\infty} \left( \theta(x^0) - \theta(x^{k+1}) \right) \leq \theta(x^0) < +\infty,$$

which implies $\lim_{i \to +\infty} \|x^{k+1} - x^i\|_{\mathcal{G}+2\mathcal{T}} = 0$. The proof of this theorem is thus complete by the positive definiteness of the operator $\mathcal{G} + 2\mathcal{T}$. $\square$

## H. Discussions on $\mathcal{G}$ and $\mathcal{T}$

Here, we discuss the roles of $\mathcal{G}$ and $\mathcal{T}$. The majorization technique used to handle the smooth function $g$ and the presence of $\mathcal{G}$ are used to make the subproblems (10) in Algorithm 3 more amenable to efficient computations. As can be observed in Theorem 6, the algorithm is convergent if $\mathcal{G} + 2\mathcal{T} \succeq 0$. This indicates that instead of adding the commonly used positive semidefinte or positive definite proximal terms, we allow $\mathcal{T}$ to be indefinite for better practical performance. Indeed, the computational benefit of using indefinite proximal terms has been observed in (Gao & Sun, 2010; Li et al., 2016a). In fact, the introduction of indefinite proximal terms in the DC algorithm is motivated by these numerical evidence. As far as we know, Theorem 6 provides the first rigorous convergence proof of the introduction of the indefinite proximal terms in the DC algorithms. The presence of $\mathcal{G}$ and $\mathcal{T}$ also helps to guarantee the existence of solutions for the subproblems (10). Since $\mathcal{G} + 2\mathcal{T} \succeq 0$ and $\mathcal{G} \succeq 0$, we have that $2\mathcal{G} + 2\mathcal{T} \succeq 0$, i.e., $\mathcal{G} + \mathcal{T} \succeq 0$ (the reverse direction holds when $\mathcal{T} \succeq 0$). Hence, $\mathcal{G} + 2\mathcal{T} \succeq 0$ ($\mathcal{G} + 2\mathcal{T} \succ 0$) implies that subproblems (10) are (strongly) convex problems. Meanwhile, the choices of $\mathcal{G}$ and $\mathcal{T}$ are very much problem dependent. The general principle is that $\mathcal{G} + \mathcal{T}$ should be as small as possible while $x^{k+1}$ is still relatively easy to compute.